\def\year{2021}\relax
\documentclass[letterpaper]{article} 
\usepackage{aaai21}  
\usepackage{times}  
\usepackage{helvet} 
\usepackage{courier}  
\usepackage[hyphens]{url}  
\usepackage{graphicx} 
\usepackage{graphicx}  
\usepackage{natbib}  
\usepackage{caption} 
\usepackage{booktabs,subcaption,amsfonts,dcolumn}
\usepackage{float}
\usepackage{comment}
\usepackage{array}
\usepackage{amsmath}

\usepackage{epstopdf}
\usepackage{multirow}
\usepackage{xcolor}
\usepackage[switch]{lineno}  %

\newcolumntype{L}[1]{>{\raggedright\let\newline\\\arraybackslash\hspace{0pt}}m{#1}}
\newcolumntype{C}[1]{>{\centering\let\newline\\\arraybackslash\hspace{0pt}}m{#1}}
\newcolumntype{R}[1]{>{\raggedleft\let\newline\\\arraybackslash\hspace{0pt}}m{#1}}

\title{CAKES: Channel-wise Automatic KErnel Shrinking for Efficient 3D Networks}
\author {
        Qihang Yu,
        Yingwei Li,
        Jieru Mei,
        Yuyin Zhou,
        Alan Yuille\\
}
\affiliations {
    Johns Hopkins University\\
    \{yucornetto, meijieru, zhouyuyiner, alan.l.yuille\}@gmail.com, 	yingwei.li@jhu.edu
}

\begin{document}


\maketitle

\begin{abstract}
3D Convolution Neural Networks (CNNs) have been widely applied to 3D scene understanding, such as video analysis and volumetric image recognition. However, 3D networks can easily lead to over-parameterization which incurs expensive computation cost.
In this paper, we propose \emph{Channel-wise Automatic KErnel Shrinking (CAKES)}, to enable efficient 3D learning by shrinking standard 3D convolutions into a set of economic operations (\emph{e.g.}, 1D, 2D convolutions). 
Unlike previous methods, CAKES performs channel-wise kernel shrinkage, which enjoys the following benefits: 1) enabling operations deployed in every layer to be heterogeneous, so that they can extract diverse and complementary information to benefit the learning process;
and 2) allowing for an efficient and flexible replacement design, which can be generalized to both spatial-temporal and volumetric data. Further, we propose a new search space based on CAKES, so that the replacement configuration can be determined automatically for simplifying 3D networks.
CAKES shows superior performance to other methods with similar model size, and it also achieves comparable performance to state-of-the-art
with much fewer parameters and computational costs on tasks including 3D medical imaging segmentation and video action recognition. Codes and models are available at \url{https://github.com/yucornetto/CAKES}.
\end{abstract}

\section{Introduction}
\label{Introduction}
3D learning has attracted more and more research attention with the recent advance of deep neural networks. However, 3D convolution layers typically result in expensive computation and suffer from convergence problems due to over-fitting issues and the lack of pre-trained weights~\cite{carreira2017quo,tajbakhsh2016convolutional}.

To resolve the redundancy in 3D convolutions, many efforts have been investigated to design efficient alternatives. For instance, \citet{qiu2017learning} and \citet{tran2018closer} propose to factorize the 3D kernel and replace the 3D convolution with Pseudo-3D (P3D) and (2+1)D convolution, where 2D and 1D convolution layers are applied in a structured manner. \citet{xie2018rethinking} suggest that replacing 3D convolutions with low-cost 2D convolutions at the bottom of the network significantly improves recognition efficiency. 

Despite their effectiveness for spatial-temporal information extraction, there are several limitations of existing alternatives to 3D convolutions. 
Firstly, these methods (\emph{e.g.}, P3D) are specifically tailored to video datasets, where data can be explicitly separated into time and space. However, for volumetric data such as CT/MRI where all three dimensions should be treated equally, conventional spatial-temporal operators can lead to biased information extraction. 
Moreover, existing operations are still insufficient even for spatial-temporal data since they may exhibit certain levels of redundancy either along the temporal or the spatial dimension, as empirically suggested in~\citet{xie2018rethinking}. Secondly, existing replacements are manually designed. Consequently, this process can be time-consuming and may lead to sub-optimal results. 

\begin{figure}[!t]
\centering
\includegraphics[width=1.0\linewidth]{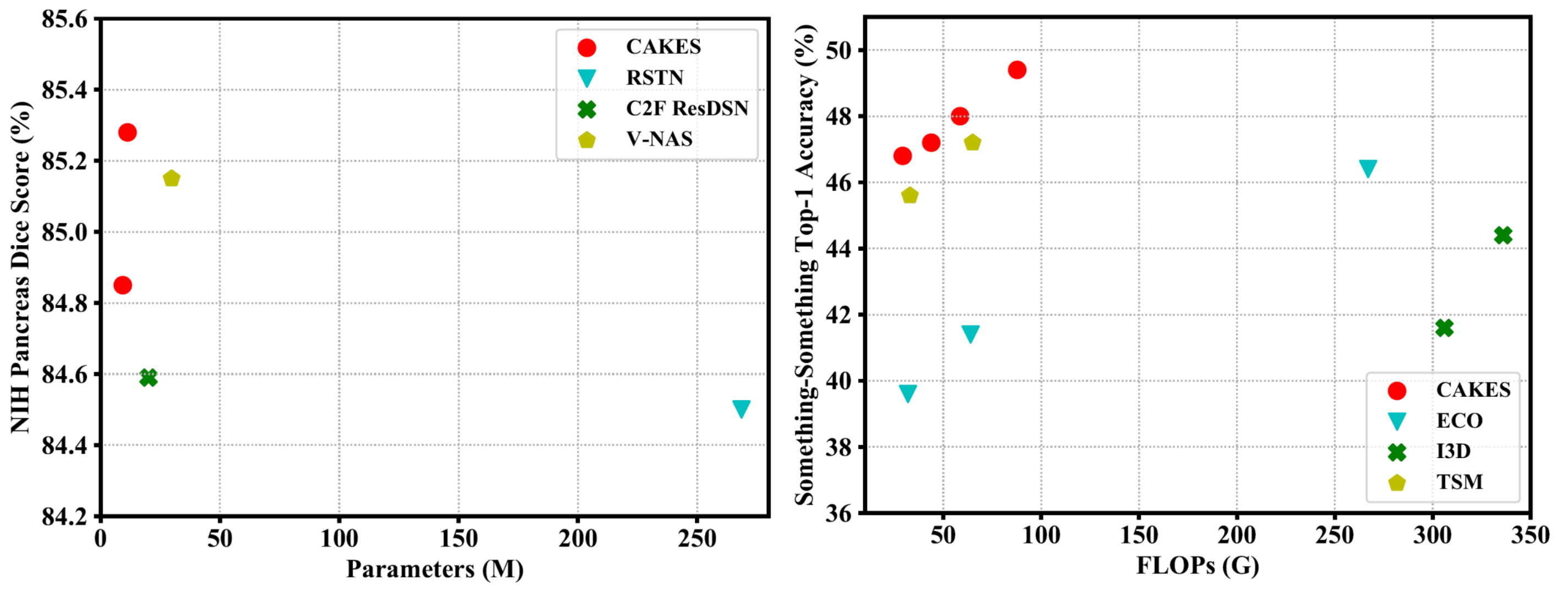}

\caption{CAKES shows better accuracy-cost trade-off on both 3D medical image segmentation (left) and action recognition (right) tasks.}
\vspace{-2ex}
\label{Fig:fig1}
\end{figure}

To address these issues, we introduce \emph{Channel-wise Automatic KErnel Shrinking (CAKES)}, as a general efficient alternatives to existing 3D operations. Specifically, the proposed method simplifies conventional 3D operations by adopting a combination of diverse and economic operations (\emph{e.g.}, 1D, 2D convolutions), where these different operators can extract complementary information to be utilized in the same layer. 
Our approach is not tailored to any specific type of input (\emph{e.g.}, videos), but can be generalized to different types of data and backbone architectures to achieve a fine-grained and efficient replacement.

As a proof test, our CAKES with a naive manual setting already exhibits superior performances compared with existing 3D replacements (Table~\ref{MSD_Table} \& \ref{STH_Table}). However, the manual selection of the set of replacing operators as well as their positioning requires trial-and-error. To further improve the performance and the model efficiency, we introduce a new search space consisting of computationally-efficient candidate operators, to facilitate the search for the optimal replacement configuration given a backbone architecture. With our search space design, the proposed CAKES is feasible to obtain a good architecture in several GPU days.

The proposed algorithm delivers high-performance and efficient models. As shown in Fig.~\ref{Fig:fig1}, evaluated on both 3D medical image segmentation and video action recognition tasks, our method achieves a better accuracy-cost trade-off. Compared with its 3D baseline, CAKES not only shows superior performance but also effectively reduces the model size (56.80\% less on medical and 19.35\% less on video) and computational cost (53.76\% less on medical and 19.01\% less on video) significantly. The proposed method surpasses their 2D/3D/P3D counterparts significantly.

Our contributions can be summarized into three folds:

(1) We propose a more generic, efficient and flexible alternative to 3D convolution by shrinking 3D kernels into heterogeneous yet complementary efficient counterparts at a fine-grained level.

(2) We automate the replacement configuration for simplifying 3D networks by customizing a search space based on CAKES and combining it with neural architecture search.

(3) By applying CAKES to different 3D models, we achieve comparable results to state-of-the-art while being much more efficient on both volumetric medical data and temporal-spatial video data.

\section{Related Work}
\label{RelatedWork}
\subsection{Efficient 3D Convolutional Neural Networks}
Despite the great advances of 3D CNNs~\cite{carreira2017quo,cciccek20163d,tran2015learning,zhou2019hyper}, 
existing 3D networks usually require heavy computational budget. Besides, 3D CNNs also suffer from unstable training due to lack of pre-trained weights~\cite{carreira2017quo,liu20173d,tajbakhsh2016convolutional}. These facts have motivated researchers to find efficient alternatives to 3D convolutions. 
For example, it is suggested in~\citet{luo2019grouped,tran2019video} to apply group convolution~\cite{krizhevsky2012imagenet} and depth-wise convolution~\cite{chollet2017xception} to 3D networks to obtain resource-efficient models. 
Another type of approach suggests replacing each 3D convolution layer with a structured combination of 2D and 1D convolution layers to achieve better performance while being more efficient. For instance, \citet{qiu2017learning} and \citet{tran2018closer} propose to use a 2D spatial convolution layer followed by a 1D temporal convolution layer to replace a standard 3D convolution layer.  Besides, \citet{xie2018rethinking} demonstrate that 3D convolutions are not needed everywhere and some of them can be replaced by 2D counterparts. Similar attempts also occur in the medical imaging area~\cite{liu20183d}. For example, \citet{gonda2018parallel} try to replace consecutive 3D convolution layers through consecutive 2D convolution layers followed by a 1D convolution layer.

Our method differs from these methods by the following folds: (1) Instead of applying homogeneous operations to all channels, CAKES allows assigning complementary heterogeneous operations at channel-wise, which leads to a more flexible design and potentially better trade-off between accuracy and efficiency~\cite{tan2019mixconv};
and (2) We enable the automatic optimization of the replacement configuration instead of manual design through a new search space.

\subsection{Neural Architecture Search}
Neural Architecture Search (NAS) aims at automatically discovering better network architectures than human-designed ones. It has been proved successful not only for 2D natural image recognition~\cite{zoph2016neural,yu2020bignas}, but also for other tasks such as segmentation~\cite{liu2019auto} and detection~\cite{ghiasi2019fpn}. Besides the success on natural images, there are also some trials on other data formats such as videos~\cite{ryoo2019assemblenet} and 3D medical images~\cite{yu2019c2fnas,zhu2019v}. Earlier NAS algorithms are based on either reinforcement learning~\cite{baker2016designing,zoph2016neural,zoph2018learning} or evolutionary algorithm~\cite{real2019regularized,xie2017genetic}. However, these methods often require training each network candidate from scratch, therefore the intensive computational costs hamper its usage especially with limited computational budget. Since~\citet{pham2018efficient} first proposed parameter sharing scheme, more and more search methods such as differentiable NAS approaches~\cite{chen2019progressive,liu2018darts,xu2019pc,dong2019searching} and one-shot NAS approaches~\cite{brock2017smash,guo2019single,stamoulis2019single,li2020neural} began to investigate how to effectively reduce the search cost to several GPU days or even several GPU hours. 

Moreover, \citet{gordon2018morphnet,mei2019atomnas} successfully connect network pruning with NAS and design more efficient search methods.
Some methods~\cite{tan2019mixconv,mei2019atomnas,stamoulis2019single} also incorporate the kernel size into the search space. Nevertheless, most of them only consider simple cases with choices among $3\times 3$, $5\times 5$, $etc.$, while we consider much more diverse and general kernel deployment across different channels in 3D settings.

\section{Method}
\label{Method}
\subsection{Revisit Variants of 3D Convolution}
\label{sec:motivation}
We first revisit 3D convolutions and existing alternatives. Without loss of generality, let $\mathrm{\textbf{X}}$ of size ${C_{i} \times D_{i} \times H_{i} \times W_{i}}$ denotes the input tensor, where $C_{i}$ stands for the input channel number, and $D_{i}$, $H_{i}$, $W_{i}$ represent the spatial depth (or temporal length), the spatial height, and the spatial width, respectively. The weights of the corresponding 3D kernel are denoted as $\mathrm{\textbf{W}}^{C_{o}\times C_{i}\times k_{d} \times k_{h} \times k_{w}}$, where $C_{o}$ is the output channel number and $k_{d} \times k_{h} \times k_{w}$ denote the kernel size. 
For simplicity, we consider each output channel individually in formulation.
Therefore, the output tensor $\mathrm{\textbf{Y}}$ of shape ${C_{o} \times D_{o} \times H_{o} \times W_{o}}$ can be derived as following:
\begin{equation}\label{eqn:regular_3D}
    \mathrm{\textbf{Y}}_{c}^{D_{o} \times H_{o} \times W_{o}} = \mathrm{\textbf{X}}^{C_{i} \times D_{i} \times H_{i} \times W_{i}}~\oplus~\mathrm{\textbf{W}}_{c}^{C_{i} \times k_{d} \times k_{h} \times k_{w}},
\end{equation} 
where $\oplus$ denotes convolution, $c$ is the output channel index, \emph{i.e.}, $1\leq c \leq C_{o}$.

The computation overhead of 3D convolutions can be significantly heavier than their 2D counterparts. Consequently, the expensive computation and over-parameterization induced by 3D deep networks impede the scalability of network capacity. Recently, there are many works seeking to alleviate the high demand of 3D convolutions.
One common strategy is to decouple the spatial and temporal components~\cite{qiu2017learning,tran2018closer}. 
The underlying assumption here is that the spatial and temporal kernels are orthogonal to each other, and therefore can effectively extract complementary information from different dimensions. 
Another option is to discard 3D convolutions and simply use 2D operations instead~\cite{xie2018rethinking}.
Mathematically speaking, these replacements can be written as:

\begin{equation}
\label{eqn:kernel_factorization}
        \mathbf{W}_{c}^{C_{i}\times k_{d}\times k_{h}\times k_{w}} \leftarrow \{\mathbf{W}_{c}^{C_{i}\times 1\times k_{h}\times k_{w}} , \mathbf{W}_{c}^{C_{i}\times k_{d}\times 1\times 1}\} 
\end{equation}
\begin{equation}
\label{eqn:2d_replacement}
    \mathbf{W}_{c}^{C_{i}\times k_{d}\times k_{h}\times k_{w}} \leftarrow \{\mathbf{W}_{c}^{C_{i}\times 1\times k_{h}\times k_{w}}\},
\end{equation}
where $\leftarrow$ indicates the replacement operation.
Similar ideas also occur in 3D medical image analysis, where the images are volumetric data. For instance, it is shown in~\citet{liu20173d} that using 2D convolutions in encoder and replacing 3D convolutions with Pseudo-3D (P3D) operations in decoder not only largely reduce the computation overhead but also improve the performance over the traditional 3D networks. 

Though these methods have furthered the model efficiency compared with standard 3D convolutions, there are several limitations yet to be tackled.
On the one hand, as shown in Eqn.~\eqref{eqn:kernel_factorization}, decomposing the kernels into orthogonal 2D and 1D components is designed for a specific data type (\emph{i.e.}, spatial-temporal), which may not well generalize to other types such as volumetric data.
On the other hand, directly replacing 3D kernels with 2D operators (Eqn.~\eqref{eqn:2d_replacement}) cannot effectively capture information along the third dimension. 

To address these issues, we propose \emph{Channel-wise Automatic KErnel Shrinking (CAKES)}, as a general alternative to 3D convolutions. The core idea is to shrink standard 3D kernels into a set of cheaper 1D, 2D, and 3D components. To ensure the flexibility of our design and avoid the tricky manual configuration, we further make the shrinkage channel-specific, thus heterogeneous kernels can extract complementary information as a 3D kernel does. We additionally introduce a brand-new search space so that the replacement configuration can be optimized automatically.

\subsection{Kernel Shrinking as Path-level Selection}
\label{sec:kernel_shrinking_formulation}

Let's consider the case for single output channel, and abbreviate $\mathbf{W}_{c}^{C_{i}\times k_{d}\times k_{h}\times k_{w}}$ to $\mathbf{W}_{c}^{k_{d} \times k_{h}\times k_{w}}$ for simplicity.
We aim to find the optimal sub-kernel $\textbf{W}_{c}^{k^{'}_{d}\times k^{'}_{h} \times k^{'}_{w}}$ ($1\leq k^{'}_{d}\leq k_{d}$,$1\leq k^{'}_{h}\leq k_{h}$,$1\leq k^{'}_{w}\leq k_{w}$) as the substitute for 3D kernel $\textbf{W}_{c}^{k_{d}\times k_{h} \times k_{w}}$. Therefore, the original 3D kernels can be effectively reduced to smaller sub-kernels, leading to a more efficient model.

\begin{figure}[!t]
\centering
\includegraphics[width=1.0\linewidth]{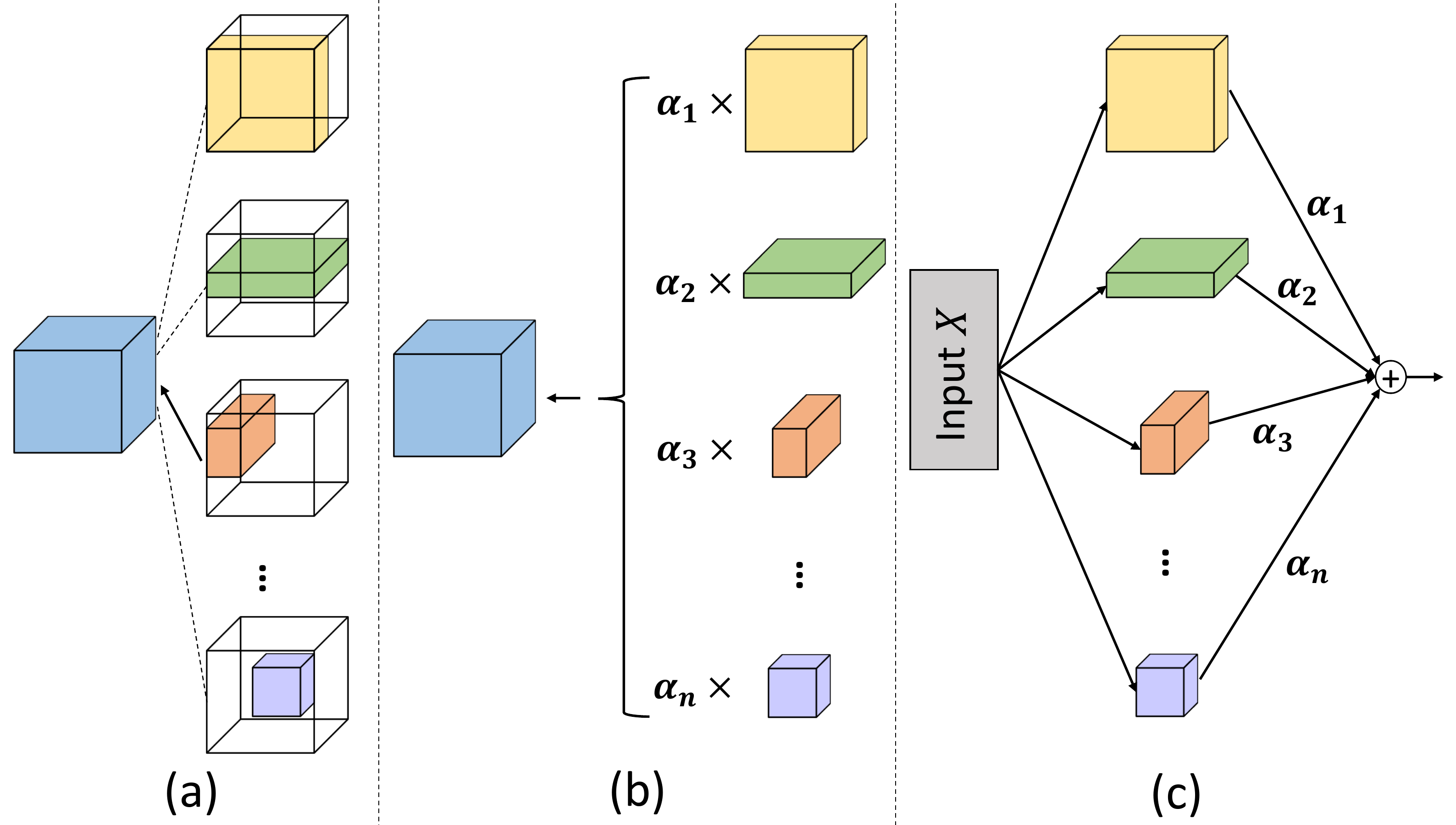}
\caption{(a) Various sub-kernels of the same 3D kernel. (b) Representation of 3D kernel as weighted summation of sub-kernels. (c) Path-level selection.}
\vspace{-2ex}
\label{Fig:Op_Select}
\end{figure}

As shown in Fig.~\ref{Fig:Op_Select}(a), even only considering different kernel sizes, there are $k_{d}\times k_{h}\times k_{w}$ sub-kernel options for a 3D kernel, which makes it impractical to find the optimal sub-kernel via manual designs. Therefore, we provide a new perspective---to formulate this problem as path-level selection~\cite{liu2018darts}, \emph{i.e.}, to encode sub-kernels into a multi-path super-network and select the optimal path among them (Fig.~\ref{Fig:Op_Select}(c)). Then this problem can be solved in a differentiable manner.

We first represent a general replacement to 3D kernel as follows (Fig.~\ref{Fig:Op_Select}(b)):
\begin{equation}
    \textbf{W}_{c}^{k_{d} \times k_{h} \times k_{w}} \leftarrow \{ \alpha_{i}\textbf{W}_{c}^{k^{i}_{d}\times k^{i}_{h} \times k^{i}_{w}}\}_{i},
\end{equation}
where $\alpha_{i}$ is the weight of $i$-th sub-kernel $\textbf{W}_{c}^{k^{i}_{d}\times k^{i}_{h} \times k^{i}_{w}}$, $1\leq k^{i}_{d}\leq k_{d}$, $1\leq k^{i}_{h}\leq k_{h}$, $1\leq k^{i}_{w}\leq k_{w}$.
With this formulation, the problem of finding the optimal sub-kernel of $\mathbf{W}_{c}^{k_{d}\times k_{h} \times k_{w}}$ can be approximated as finding the optimal setting of $\{\alpha_{i}\}$ and then keeping the sub-kernel with maximum $\alpha_i$.
Due to the linearity of convolution, Eqn.~\eqref{eqn:regular_3D} can then be derived as below:
\begin{equation}\label{eqn:multi_path}
    \textbf{X} \oplus \textbf{W}_{c}^{k_{d} \times k_{h} \times k_{w}} \leftarrow \sum_{i} \alpha_{i}(\textbf{X} \oplus \textbf{W}_{c}^{k^{i}_{d} \times k^{i}_{h} \times k^{i}_{w}}).
\end{equation}

To solve for the path weights $\{\alpha_{i}\}$, we reformulate Eqn.~\eqref{eqn:multi_path} as an over-parameterized multi-path super-network, where each candidate path consists of a sub-kernel (Fig.~\ref{Fig:Op_Select}(c)). By relaxing the selection space, $\emph{i.e.}$, relaxing the conditions on $\alpha$ to be continuous, Eqn.~\eqref{eqn:multi_path} can be then formulated as a differential NAS problem and optimized via gradient descent~\cite{liu2018darts}.

\begin{figure}[!t]
\centering
\includegraphics[width=1.0\linewidth]{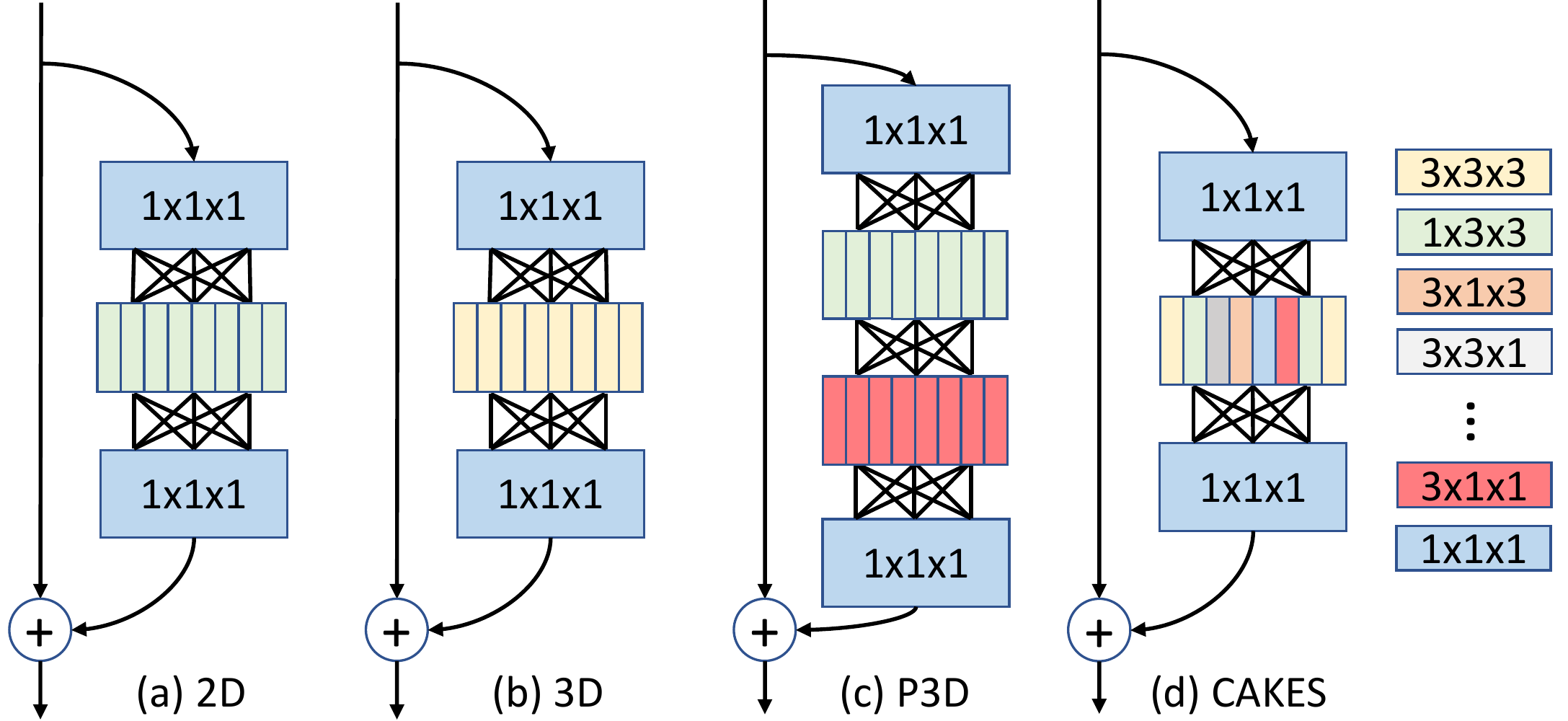}
\vspace{-2ex}
\caption{An illustrative example of comparison between different types of convolution in a residual block~\cite{he2016deep}. (a) 2D Convolution. (b) 3D Convolution. (c) P3D Convolution. (d) the proposed CAKES. In our case, starting from a 3D convolution, the 3D operation at each channel is replaced with an efficient sub-kernel.}
\vspace{-2ex}
\label{Fig:Conv_Compare}
\end{figure}
\subsection{Channel-wise Shrinkage}
\label{sec:channel_wise_shrinkage}
While previous replacements~\cite{liu20173d,qiu2017learning,tran2018closer} consist of homogeneous operations in the same layer, we argue that a more efficient replacement requires customized operations at each channel.
As shown in Fig.~\ref{Fig:Conv_Compare}, kernel shrinking in a channel-wise fashion can generate heterogeneous operations which extract diverse and complementary information within the same layer, and thereby yields a fine-grained and more efficient replacement (Fig.~\ref{Fig:Conv_Compare}(d)) than prior methods which use layer-wise replacements (Fig.~\ref{Fig:Conv_Compare}(a) \& (b) \& (c)).

Contrary to previous layer-wise replacement, our core idea is to replace 3D kernel at each channel individually, thus the target is to find the optimal sub-kernel $\mathbf{W}_{c}^{k^{c}_{d}\times k^{c}_{h}\times k^{c}_{w}}$ as the substitute for the $c$-th output channel 3D kernel $\mathbf{W}_{c}^{k_{d}\times k_{h}\times k_{w}}$:
\begin{equation}
\mathbf{W}_{c}^{k_{d}\times k_{h}\times k_{w}} \leftarrow \{\mathbf{W}_{c}^{k^{c}_{d}\times k^{c}_{h}\times k^{c}_{w}}\},
\end{equation}
where the optimal size of the sub-kernel ($k^{c}_{d} \times k^{c}_{h} \times k^{c}_{w}$) is subjected to $1\leq k^{c}_{d}\leq k_{d}$, $1\leq k^{c}_{h}\leq k_{h}$, $1\leq k^{c}_{w}\leq k_{w}$. Hence the computation incurred by Eqn.~\eqref{eqn:regular_3D} can be largely reduced by our replacement as above.

With our channel-wise replacement design, the original 3D kernels are substituted by a series of diverse and cheaper operations at different channels as following (recall that $C_{o}$ is the output channel number):
\begin{equation}
\small
    \mathbf{W} \leftarrow \{\mathbf{W}_{1}^{k^{1}_{d}\times k^{1}_{h}\times k^{1}_{w}}, \mathbf{W}_{2}^{k^{2}_{d}\times k^{2}_{h}\times k^{2}_{w}},\dots,\mathbf{W}_{C_{o}}^{k^{C_{o}}_{d}\times k^{C_{o}}_{h}\times k^{C_{o}}_{w}}\}.
\end{equation}

Benefited from channel-wise shrinkage, our method provides a more general and flexible design for replacing 3D convolution than previous approaches (Eqn.~\eqref{eqn:kernel_factorization} and Eqn.~\eqref{eqn:2d_replacement}), where it can also be easily reduced to arbitrary alternatives ($e.g.$, 2D, P3D) by integrating these operations into the set of candidate sub-kernels. An illustration example can be found in Fig~\ref{Fig:Conv_Compare}.
\subsection{Search for an Efficient Replacement}
\label{sec:NAS}
As aforementioned, given the tremendous feasible choices, it is impractical to manually find the optimal replacement for a 3D kernel through a trial-and-error process. 
Especially, it becomes even more intractable as the replacement procedure is conducted in a channel-wise manner.
Therefore, we propose a new search space for efficient 3D networks and automate the process of learning an efficient replacement to fully exploit the redundancies in 3D convolution operations. 
By formulating kernel shrinkage as a path-level selection problem, we first construct a super-network where every candidate sub-kernel is encapsulated into a separate trainable branch (Fig.~\ref{Fig:Op_Select}(c)) at each channel. 
Once the path weights are learned in a differentiable manner, the optimal path (sub-kernel) can be determined.

\vspace{1.0ex}
\noindent\textbf{Search Space.}
A well-designed search space is crucial for NAS algorithms~\cite{yang2019evaluation}. Here we aim to answer the following questions: \emph{Should the 3D convolution kernel be kept or replaced per channel? If replaced, which operation should be deployed instead?}

To address these questions, for each channel, we define a set $\mathcal{S}$, which contains all candidates of sub-kernels (replacement) given a 3D kernel $\textbf{W}^{k_d \times k_{h}\times k_{w}}$:
\begin{equation}
\small
\begin{gathered}
    \mathcal{S} = \{\mathbf{W}^{k_{d_{1}}\times k_{h_{1}}\times k_{w_{1}}}, \mathbf{W}^{k_{d_{2}}\times k_{h_{2}}\times k_{w_{2}}},\dots,\mathbf{W}^{k_{d_{n}}\times k_{h_{n}}\times k_{w_{n}}}\} \\
    \mathbf{W}^{k^{c}_{d}\times k^{c}_{h}\times k^{c}_{w}}_{c}=~\mathrm{Choose}(\mathcal{S}).
\end{gathered}
\end{equation}
As the original 3D convolution kernel can be considered sub-kernel of itself, \textit{i.e.}, $\textbf{W}^{k_d \times k_{h}\times k_{w}} \in \mathcal{S}$ , it can be kept in the final configuration. The final optimal operation $\mathbf{W}_{c}$ is chosen among $\mathcal{S}$.

Another critical problem for NAS is how to reduce the search cost. To make the search cost affordable, we adopt a differentiable NAS paradigm where the model structure is discovered in a single-pass super-network training.
Drawing inspirations from previous NAS methods, we directly use the scaling parameters in the normalization layers as the path weights $\alpha$ of the multi-path super-network (Eqn.~\eqref{eqn:multi_path})~\cite{gordon2018morphnet,mei2019atomnas}. 
And our goal is then equivalent to finding the optimal sub-network architecture based on the learned path weights.
To achieve this goal, we introduce two different search manners which aim at either maximizing the performance or optimizing the computation cost of the sub-network as a search priority, named as performance-priority and cost-priority search, respectively.

\vspace{1.0ex}
\noindent\textbf{Performance-Priority Search.} 
The search aims to maximize the performance by finding the optimal sub-kernels given the backbone architecture. 
During the search procedure, following~\citet{bender2018understanding,brock2017smash}, we randomly pick an operation for each channel at each iteration. This not only allows for memory saving by activating and updating one path per iteration but also propels the weights of the
paths in the super-network training to be decoupled.
After the super-network is trained, the operation with the largest path weight will be picked as the final choice for the given output channel:
\begin{equation}
\begin{gathered}
    \textbf{W}_{c}^{k_{d}\times k_{h}\times k_{w}} \leftarrow \{\textbf{W}^{k_{d_{i^{*}}}\times k_{h_{i^{*}}}\times k_{w_{i^{*}}}}\},\\ ~\textup{where}\enspace i^{*} = \mathrm{argmax}_{i \in {\{1\cdots n\}}}(\alpha_{i}).
\end{gathered}
\end{equation}

\noindent\textbf{Cost-Priority Search.}
Performance-priority may neglect the possible negative effects on the computation cost. 
In order to obtain more compact models, we also introduce a ``cost-priority" search method.
Inspired by~\citet{mei2019atomnas}, we search the model in a pruning manner with penalty on expensive operations. The outputs of each sub-kernels are concatenated and aggregated by the following $1\times1\times1$ convolution.
To make the searched architecture more compact, we introduce a ``cost-aware" penalty term---A lasso term on $\alpha$ which is used as the penalty loss to push many path weights to near-zero values. Therefore, the total training loss $\mathcal{L}$ can be written as:
\begin{equation}
    \mathcal{L} = \mathcal{E} + \lambda\sum_{i}\beta_{i}|\alpha_{i}|,
\end{equation}
where $\beta_{i}$ is a ``cost-aware" term to balance the penalty term, which is proportional to the parameters or FLOPs cost of the sub-kernel.
In Table~\ref{MSD_Table}, we also empirically show that this term can lead to a more efficient architecture. The introduction of $\beta_{i}$ aims at giving more penalty to ``expensive" operations and leading to a more efficient replacement. $\lambda$ is the coefficient of the penalty term, and $\mathcal{E}$ is the conventional training loss ($e.g.$, cross-entropy loss combined with the regularization term such as weight decay).

\section{Experiments}
\label{Experiments}

\subsection{3D Medical Image Segmentation}

\noindent\textbf{Dataset.}
We evaluate the proposed method on two public datasets: 1) Pancreas Tumours dataset from the Medical Segmentation Decathlon Challenge (MSD)~\cite{simpson2019large}, which contains 282 cases with both pancreatic tumours and normal pancreas annotations; and 2) NIH Pancreas Segmentation dataset~\cite{roth2015deeporgan}, consisting of 82 abdominal CT volumes. 
\textbf{For the MSD dataset}, we use 226 cases for training and evaluate the segmentation performance on the
rest 56 cases. The resolution along the axial axis of this dataset is extremely low and the number of slices can be as small as 37. For data preprocessing, all images are resampled
to an isotropic 1.0 $mm^3$ resolution. \textbf{For the NIH dataset}, the resolution of each scan is $512\times 512\times L$, where $L\in [181,466]$ is the number of slices along the axial axis and the voxel spacing ranges from 0.5 $mm$ to 1.0 $mm$. We test the model in a 4-fold cross-validation manner following previous methods~\cite{zhou2017fixed,zhou20192d}.

\begin{table}[tb]
\setlength{\tabcolsep}{2.0pt}
\centering
\scriptsize
\begin{tabular}{L{0.20\linewidth}C{0.11\linewidth}C{0.11\linewidth}C{0.11\linewidth}C{0.11\linewidth}C{0.11\linewidth}}
\toprule
Methods         & Params (M) & FLOPs (G) & Pancreas DSC (\%) & Tumor DSC (\%) & Average DSC (\%) \\ \midrule
$\mathrm{2D}$     & 11.29          & 97.77    & 79.16    & 43.02 & 61.09   \\
$\mathrm{3D}$     & 22.50          & 188.48    & 80.34    & 47.57 & 63.96   \\
$\mathrm{P3D}$    & 13.16          & 112.88    & $\mathbf{80.36}$    & 45.27 & 62.82   \\ \midrule
$\mathrm{CAKES^{M}_{1D}}$  & 7.56          & 67.53     & 79.77    & 42.73 & 61.25   \\
$\mathrm{CAKES^{M}_{2D}}$  & 11.29          & 97.77     & 80.09    & 46.17 & 63.13   \\
$\mathrm{CAKES^{M}_{1,2,3D}}$ & 11.41          & 99.17     & 79.82    & 45.27 & 62.55   \\ \midrule
$\mathrm{CAKES^{P}_{1D}}$  & $\mathbf{7.56}$          & $\mathbf{67.53}$     & 80.32    & 45.57 & 62.95   \\
$\mathrm{CAKES^{P}_{2D}}$  & 11.29          & 97.77     & 80.05    & 48.51 & 64.28   \\
$\mathrm{CAKES^{P}_{1,2,3D}}$ & 11.26          & 99.68     & 80.12    & $\mathbf{48.72}$ & $\mathbf{64.42}$   \\
$\mathrm{CAKES^{C}_{1,2,3D}}$ & 9.72         & 87.16     & 80.34    & 47.95 & 64.15   \\ \bottomrule
\end{tabular}
\caption{Comparison among different operations and configurations. 
The subscripts of 1D, 2D, and 3D indicate the dimensions of the operations being used. The superscripts of ``M", ``P", ``C" represent ``Manual", ``Performance-Priority", and ``Cost-Priority" respectively.
}
\label{MSD_Table}
\end{table}

\vspace{1.0ex}
\noindent\textbf{Implementation Details.} 
For all experiments, C2FNAS~\cite{yu2019c2fnas} is used as the backbone architecture. 
When replacing the operations, we keep the stem (the first two and the last two convolution layers) as the same. For 3D medical image, for simplicity, we choose a set of most representative sub-kernels as $\mathcal{S}$. The operations set contains conv1D ($1\times1\times3$, $1\times3\times1$, $3\times1\times1$), conv2D ($1\times3\times3$, $3\times1\times3$, $3\times3\times1$) from different directions, and conv3D ($3\times3\times3$). 
For every 3D kernel at each output channel, a sub-kernel from $\mathcal{S}$ will be chosen as the replacement.
\textbf{For manual settings}, we assign all candidates operations uniformly across the output channels.
\textbf{For NAS settings}, we include both ``performance-priority'' and ``cost-priority'' search for performance comparison.

\vspace{1.0ex}
\noindent\textbf{Training stage.} For the MSD dataset, we use random crop with patch size of $96\times96\times96$, random rotation ($0^{\circ}$, $90^{\circ}$, $180^{\circ}$, and $270^{\circ}$) and flip in all three axes as data augmentation. The batch size is 8 with 4 GPUs. We use SGD optimizer with learning rate starting from 0.01 with polynomial decay of power of 0.9, momentum of 0.9, and weight decay of 0.00004. The training lasts for 40k iters. The loss function is the summation of dice loss~\cite{milletari2016v} and cross-entropy loss. For NIH dataset, the patch size is set as $96\times 96\times 64$, following the settings in~\citet{zhu2019v}. The found architecture will be trained \textbf{from scratch} to ensure fair comparison. Both the super-network and the found architecture are trained under the same settings as aforementioned. For search stage with ``cost-priority" setting, a lasso term with coefficient $\lambda=1.0\times 10^{-4}$ is applied to the path weights. And it is further re-weighted by $\beta = $ $\{\frac{9}{13}, \frac{3}{13}, \frac{1}{13}\}$ for 3D, 2D, 1D operations respectively, which is their ratio of the parameters. After the training process, the operation with the largest $\alpha$ is chosen as the final replacement for 3D operation for each channel.

\vspace{1.0ex}
\noindent\textbf{Testing stage.} We test the network in a sliding-window manner, where the patch size is $96\times96\times96$ and stride is $32\times32\times32$ for the MSD dataset and patch size is $96\times96\times64$ and stride is $20\times20\times20$ for NIH dataset. The result is measured with Dice-S\o rensen coefficient (DSC) metric, which is formulated as $\mathrm{DSC}$ $(\mathcal{Y},\mathcal{Z}) = \frac{2\times|\mathcal{Y}\cap \mathcal{Z}|}{|\mathcal{Y}|+|\mathcal{Z}|}$, where $\mathcal{Y}$ and $\mathcal{Z}$ denote the prediction and ground-truth voxels set for a foreground class. The DSC has a range of $[0, 1]$ with 1 implying a perfect prediction.

\begin{table}[tb]
\setlength{\tabcolsep}{2.0pt}
\centering
\scriptsize
\begin{tabular}{L{0.40\linewidth}C{0.12\linewidth}C{0.12\linewidth}C{0.12\linewidth}C{0.12\linewidth}}
\toprule
Method & Params & Average DSC & Max DSC & Min DSC \\ \midrule

Coarse-to-fine~\cite{zhou2017fixed}   & 268.56M   & 82.37\%           & 90.85\%       & 62.43\%       \\
RSTN~\cite{yu2018recurrent}   & 268.56M  & 84.50\%           & 91.02\%       & 62.81\%       \\
C2F ResDSN~\cite{zhu20173d}   & 20.06M  & 84.59\%           & 91.45\%       & 69.62\%       \\
V-NAS~\cite{zhu2019v}  & 29.74M & 85.15\%     & 91.18\% & 70.37\% \\ \midrule
$\mathrm{CAKES^{C}_{1,2,3D}}$   & $\mathbf{9.27M}$   & 84.85\%  & 91.61\% & 59.32\% \\
$\mathrm{CAKES^{P}_{1,2,3D}}$   & 11.26M   & $\mathbf{85.28\%}$  & $\mathbf{91.98\%}$ & $\mathbf{72.78\%}$ \\ \bottomrule
\end{tabular}
\caption{Comparison with prior arts on the NIH dataset.
}
\label{NIH_Compare}
\end{table}

\vspace{1.0ex}
\noindent\textbf{Manual Settings vs. Auto Settings.} 
As observed from Table~\ref{MSD_Table}, even under manual settings, CAKES is already much more efficient with slightly inferior performance ($e.g.$, from $\mathrm{3D}$ to manual $\mathrm{CAKES^{M}_{2D}}$, parameters drop from 22.50M to 11.29M, and FLOPs drop from 188.48G to 97.77G, with performance gap of $<$ 1.0\%). Besides, $\mathrm{CAKES^{M}_{2D}}$ outperforms its counterpart with standard convolution 2D layers by more than $2.0\%$ with the same model size, which indicates the benefits of our design. In addition, with the proposed search space and method, $\mathrm{CAKES}$ can further reduce the performance gap and even surpasses the original 3D model with much fewer parameters and computations, \emph{e.g.}, model size is reduced from 22.50M ($\mathrm{3D}$) to 11.26M ($\mathrm{CAKES^{P}_{1,2,3D}}$), and FLOPs drop from 188.48G ($\mathrm{3D}$) to 99.68G ($\mathrm{CAKES^{P}_{1,2,3D}}$), with a performance improvement of 0.46\%. Compared with $\mathrm{P3D}$, $\mathrm{CAKES^{P}_{1,2,3D}}$ also yields superior performance (+1.60\%) with a more compact model (11.26M vs. 13.16M), which further indicates the effectiveness of the proposed method.

\vspace{1.0ex}
\noindent\textbf{Influence of the Search Space.}
From Table~\ref{MSD_Table}, we can see that using different search space, CAKES consistently outperforms its counterparts with standard 1D/2D/3D convolutions.
Out of different search spaces, we find that $\mathrm{CAKES^{P}_{1D}}$ (7.56M params and 67.53G FLOPs) offers the most efficient model with a comparable performance, while $\mathrm{CAKES^{P}_{2D}}$ (11.29M params and 97.77G FLOPs) can already surpass the 3D baseline (22.50M params and 188.48G FLOPs) with half parameters and computation cost. After we enlarge the search space, $\mathrm{CAKES^{P/C}_{1,2,3D}}$ obtains a configuration with even higher performance/efficiency (last 2 rows of Table~\ref{MSD_Table}).

\vspace{1.0ex}
\noindent\textbf{Generalization to different backbone architectures.} We also test our method on different backbone architectures. Applying $\mathrm{CAKES^{C}_{1,2,3D}}$ to another strong model 3D ResDSN~\cite{zhu20173d,li2019volumetric}, our method consistently leads to a more efficient model with much fewer parameters (10.03M to 4.63M) and FLOPs (192.07G to 98.12G) with comparable performance (61.96\% to 61.65\%).

\vspace{1.0ex}
\noindent\textbf{NIH Results.} We compare CAKES with state-of-the-art methods in Table~\ref{NIH_Compare}, where it can be observed that the proposed method leads to a much more compact model size compared to other models. For instance, our model size is more than $25\times$ smaller than that of~\citet{zhou2017fixed} and~\citet{yu2018recurrent}.
It is well worth noting that our model performed in a single-stage fashion already outperforms many state-of-the-art methods conducted in a two-stage coarse-to-fine manner~\cite{zhou2017fixed,yu2018recurrent,zhu20173d} on the NIH pancreas dataset with much fewer model parameters and FLOPS.
It is also noteworthy to mention that the applied architecture is searched from another dataset (MSD), where images are collected under different protocols and have different resolutions. This result indicates the generalization of our searched model.
By directly applying the architecture searched on the MSD dataset, our method also outperforms~\citet{zhu2019v} which was directly searched on the NIH dataset with less than half model size.

\setlength{\tabcolsep}{8pt}
\begin{table}[tb]
\centering
\scriptsize
\begin{tabular}{L{0.22\linewidth}C{0.09\linewidth}C{0.09\linewidth}cc}
\toprule
Model  & Params (M) & FLOPs (G) & top1 & top5\\ \midrule
C2D    & 23.9          & 33.0     & 17.2                                             & 43.1\\ 
P3D     & 27.6          & 37.9     & 44.8                                             & 74.6\\
C3D    & 46.5          & 62.6     & 46.8                                             & 75.3\\ 
\midrule
$\mathrm{CAKES^{M}_{1,2D}}$ & $\mathbf{20.1}$          & $\mathbf{28.0}$     & 46.2                                             & 75.2\\ 
$\mathrm{CAKES^{M}_{2,3D}}$ & 35.2          & 47.7     & 46.8                                             & 76.0\\ 
\midrule
$\mathrm{CAKES^{P}_{1,2D}}$   & 20.9          & 29.1     & 47.1                                             & 75.9\\ 
$\mathrm{CAKES^{P}_{2,3D}}$  & 37.5          & 50.7     & $\mathbf{47.4}$                                             & $\mathbf{76.1}$\\ 
$\mathrm{CAKES^{P}_{1,2,3D}}$  & 33.5          & 43.9     & 47.2                                             & 75.7\\ 
$\mathrm{CAKES^{C}_{1,2D}}$   & 20.5          & 29.3     & 46.8                                             & 76.0\\ 
$\mathrm{CAKES^{C}_{2,3D}}$   & 35.7          & 41.4     & 46.9                                             & 75.6\\ 
$\mathrm{CAKES^{C}_{1,2,3D}}$   & 35.0          & 38.7     & 46.9                                             & 75.5\\
\bottomrule
\end{tabular}
\caption{Comparison among operations and configurations for ResNet50 backbone in terms of parameter number, computation amount (FLOPs), and performance on Something-Something V1 dataset. 
}
\label{STH_Table}
\end{table}
\setlength{\tabcolsep}{1.4pt}

\setlength{\tabcolsep}{10pt}
\begin{table*}[t]
\label{tab:compare_something}
\begin{center}
\scriptsize
\begin{tabular}{ccccccc}
\toprule
\textbf{Method}
& \textbf{Backbone Architecture} & \textbf{\#Frame} & \textbf{FLOPs} & \textbf{\#Param.} & \textbf{top1} & \textbf{top5}\\ 
\midrule
TSN~\cite{wang2016temporal} & ResNet-50 & 8  & 33G & 24.3M & 19.7 & 46.6 \\ 
TRN-2stream~\cite{zhou2018temporal} & BNInception & 8+8 & - & 36.6M & 42.0 & - \\ \midrule
ECO~\cite{zolfaghari2018eco} & BNIncep+3D Res18  & 8  & 32G & 47.5M & 39.6 & -\\
ECO~\cite{zolfaghari2018eco} & BNIncep+3D Res18 & 16 & 64G & 47.5M & 41.4 & -\\
$\mathrm{ECO}_{En}Lite$~\cite{zolfaghari2018eco} & BNIncep+3D Res18 & 92 & 267G & 150M & 46.4 & -\\\midrule
I3D~\cite{carreira2017quo} & 3D ResNet-50 & 32$\times$2clip & 153G$\times$2  & 28.0M & 41.6 & 72.2\\ 
NL I3D~\cite{wang2018non} & 3D ResNet-50 & 32$\times$2clip & 168G$\times$2 & 35.3M & 44.4 & 76.0\\  
NL I3D+GCN~\cite{wang2018videos} & 3D ResNet-50+GCN & 32$\times$2clip & 303G$\times$2 & 62.2M & 46.1 & 76.8\\  \midrule
TSM~\cite{lin2019tsm} & ResNet-50 & 8 & 33G & 24.3M & 45.6 & 74.2 \\ 
TSM~\cite{lin2019tsm} & ResNet-50 & 16 & 65G & 24.3M & 47.2 & 77.1  \\\midrule
S3D~\cite{xie2018rethinking} & BNInception & 64 & 66.38G & - & 47.3 & 78.1\\ 
S3D-G~\cite{xie2018rethinking} & BNInception & 64 & 71.38G & - & 48.2 & \textbf{78.7}\\ 
\midrule
$\mathrm{CAKES^{C}_{1,2D}}$ & ResNet-50 & 8 & \textbf{29.3G} & \textbf{20.5M} & 46.8 & 76.0\\
$\mathrm{CAKES^{P}_{2,3D}}$ & ResNet-50 & 8 & 50.7G & 37.5M & \textbf{47.4} & 76.1\\
$\mathrm{CAKES^{P}_{1,2,3D}}$ & ResNet-50 & 8 & 43.9G & 33.5M & 47.2 & 75.7\\ \midrule

$\mathrm{CAKES^{C}_{1,2D}}$ & ResNet-50 & 16 & \textbf{58.6G} & \textbf{20.5M} & 48.0 & 78.0\\
$\mathrm{CAKES^{P}_{2,3D}}$ & ResNet-50 & 16 & 101.4G & 37.5M & 48.6 & 78.6\\
$\mathrm{CAKES^{P}_{1,2,3D}}$ & ResNet-50 & 16 & 87.8G & 33.5M & \textbf{49.4} & 78.4\\
\bottomrule
\end{tabular}
\caption{Comparing CAKES against other methods on Something-Something V1 dataset. We mainly consider the methods that adopt convolutions in fully-connected manner and only take RGB as input for fair comparison.}
\label{sth_comapre_table}
\end{center}
\end{table*}

\subsection{Action Recognition in Videos}
\noindent\textbf{Dataset.}
Something-Something V1~\cite{goyal2017something} is a large scale action recognition dataset which requires comprehensive temporal modeling. There are totally about 110k videos for 174 classes with diverse objects, backgrounds, and viewpoints. 

\vspace{1.0ex}
\noindent\textbf{Implementation Details.}
We adopt ResNet50~\cite{he2016deep} with pre-trained weight on ImageNet~\cite{krizhevsky2012imagenet} as our backbone. The 3D convolution weights are initialized by repeating 2D kernel by 3 times along the temporal dimension following~\cite{carreira2017quo}, while 1D convolution weights are initialized by averaging the 2D kernel on spatial dimensions and then repeat by 3 times along temporal axis. For the temporal dimension, we use the sparse sampling method as in~\cite{wang2016temporal}. For spatial dimension, the short side of the input frames are resized to 256 and then cropped to $224\times224$.

\vspace{1.0ex}
\noindent\textbf{Training Stage.} We use random cropping and flipping as data augmentation. We train the network with a batch size of 96 on 8 GPUs with SGD optimizer. The learning rate starts from 0.04 for the first 50 epochs and decays by a factor of 10 for every 10 epochs afterwards. The total training epochs are 70. We also set dropout ratio to 0.3 following~\cite{wang2018videos}. The training settings remain the same for both final network and search stage, except that when searching with ``performance-priority" we double the training epochs to ensure convergence, and with ``cost-priority", we use a lasso term with $\lambda=1.0\times 10^{-4}$ and $\beta=$ $\{\frac{9}{13}, \frac{3}{13}, \frac{1}{13}\}$ for 3D, 2D, 1D operations respectively. 

\vspace{1.0ex}
\noindent\textbf{Testing Stage.} we sample the middle frame in each segment and perform center crop for each frame. We report the results of \textbf{single crop}, unless otherwise specified.

\vspace{1.0ex}
\noindent\textbf{Ablation Study.} We study the impacts of both different operations set and manual/auto configurations. The results are summarized in Table~\ref{STH_Table}. Considering the spatial-temporal property of video data, we study the following different operations set: (1) Spatial 2D convolution and temporal 1D convolution; (2) Spatial 2D convolution and 3D convolution; (3) Spatial 2D, temporal 1D, and 3D convolutions. 

\vspace{1.0ex}
\noindent\textbf{Operation Set with 1D \& 2D Sub-kernels.}
As shown in Table~\ref{STH_Table}, $\mathrm{CAKES^{C}_{1,2D}}$ surpass the 2D baseline by a large margin (+29.6\%)
, while the model size reduces by 14.23\%.
This suggests that TSN~\cite{wang2016temporal} may lack the ability to capture temporal information, therefore replacing some of the 2D operations to temporal 1D operations can significantly increase the performance and reduce the model size. 
Besides, it also surpasses P3D, where each 2D convolution is followed by a temporal 1D convolution, with a significant advantage on both performance (+2.0\%)
and model cost (25.72\% fewer params and 53.19\% fewer FLOPs),
indicating $\mathrm{CAKES}$ makes better use of redundancies in the networks than P3D. Therefore, $\mathrm{CAKES}$ using operation set containing 1D and 2D sub-kernels can be an ideal design when looking for efficient video understanding networks.

\vspace{1.0ex}
\noindent\textbf{Operation Set with 2D \& 3D Sub-kernels.}
We aim to see how $\mathrm{CAKES}$ balances the trade-off between performance and model cost. From Table~\ref{STH_Table}, $\mathrm{CAKES^{C}_{2,3D}}$ yields a much more compact model
(-23.23\%/33.87\% params/FLOPs)
with a comparable performance to C3D. Under ``performance-priority" setting, $\mathrm{CAKES^{P}_{2,3D}}$ searches a slightly larger model
, yet its performance boosts significantly to 47.4\%. 

\vspace{1.0ex}
\noindent\textbf{Operation Set with 1D \& 2D \& 3D Sub-kernels.}
Compared to $\mathrm{CAKES^{C}_{2,3D}}$, $\mathrm{CAKES^{C}_{1,2,3D}}$ shows a similar performance with much fewer FLOPs (38.7G vs. 41.4G). Besides, under the ``performance-priority'' setting, $\mathrm{CAKES^{P}_{1,2,3D}}$ produces a comparable performance to $\mathrm{CAKES^{P}_{2,3D}}$ with less computation cost (43.9G vs. 50.7G). This result indicates that with a more general search space ($e.g.$, 1D, 2D, and 3D), the proposed CAKES can find more flexible designs, which lead to better performance/efficiency.

\vspace{1.0ex}
\noindent\textbf{Results.} A comparison with other state-of-the-art methods is shown in Table~\ref{sth_comapre_table}. We report the model performance under both 8-frame and 16-frame settings. Compared with other state-of-the-art methods, $\mathrm{CAKES^{P}_{2D,3D}}$ sampling only 8 frames can already outperform most current methods. With smaller parameters and FLOPs, $\mathrm{CAKES^{P}_{2D,3D}}$ surpasses those complex models such as non-local networks~\cite{wang2018non} with graph convolution~\cite{wang2018videos}. Comparing $\mathrm{CAKES^{C}_{1,2D}}$ to other efficient video understanding framework such as ECO~\cite{zolfaghari2018eco} and TSM~\cite{lin2019tsm}, our model is not only more light-weight (58.6G vs. 64G/65G), but also delivers a better performance (48.0\% vs. 41.4\%/47.2\%). And our best model $\mathrm{CAKES^{P}_{1,2,3D}}$ achieves a new state-of-the-art performance of 49.4\% top-1 accuracy with a moderate model size. An interesting finding is that although $\mathrm{CAKES^{P}_{1,2,3D}}$ shows similar performances to $\mathrm{CAKES^{P}_{2,3D}}$ with 8-frame inputs, it achieves a much higher accuracy when it comes to the 16-frame scenario, which demonstrates that with a more general search space, $\mathrm{CAKES^{P}_{1,2,3D}}$ shows stronger transferability than other counterparts.

\section{Conclusions}
\label{Conclusions}
As an important solution to various 3D vision applications, 3D networks still suffer from over-parameterization and heavy computations. How to design efficient alternatives to 3D operations remains an open problem. 
In this paper, we propose Channel-wise Automatic KErnel Shrinking (CAKES), where standard 3D convolution kernels are shrunk into efficient sub-kernels at channel-level, to obtain efficient 3D models. Besides, by formulating kernel shrinkage as a path-level selection problem, our method can automatically explore the redundancies in 3D convolutions and optimize the replacement configuration. By applying on different backbone models, the proposed CAKES significantly outperforms previous 2D/3D/P3D and other state-of-the-art methods on both 3D medical image segmentation and action recognition from videos. 

\vspace{1ex}
\noindent \textbf{Acknowledgment.} This work was supported by the Lustgarten Foundation for Pancreatic Cancer Research. We thank Zhuotun Zhu, Yingda Xia, Wei Shen and Yan Wang for instructive discussions.

\section{Ethics Statement}
In this paper, we present a new operation as an efficient alternative to 3D convolution and a search space to automate the process of simplifying 3D networks.
The findings described in this paper can potentially help reduce the computation burden especially when dealing with 3D data such as CT scan or video. For the research community, our finding sheds light on a new direction to design efficient 3D networks. We expect the methodology to be investigated further in the future to better understand and make full use of the redundancy in 3D networks. To the society, some applications can be greatly benefited from CAKES. For example, in medical applications where artificial intelligence is expected to help doctor to make diagnosis and treatment planning, model efficiency can be crucial for model deployment in real-world clinical flows. And it can also assist deploy the algorithm (\emph{e.g.} on-device video recognition) to mobile devices which do not have strong computation ability.

However, we also note that there is a long-lasting debate on the impacts of AI on human world. As a method improving the fundamental ability of deep learning, our work also advances the development of AI, which means there could be both beneficial and harmful influences depending on the users.

\bibliography{ref}


\clearpage
\onecolumn
\appendix

\section*{Appendix}
\begin{figure*}[h]
    \begin{subfigure}[b]{\textwidth}
        \centering
        \includegraphics[width=\textwidth]{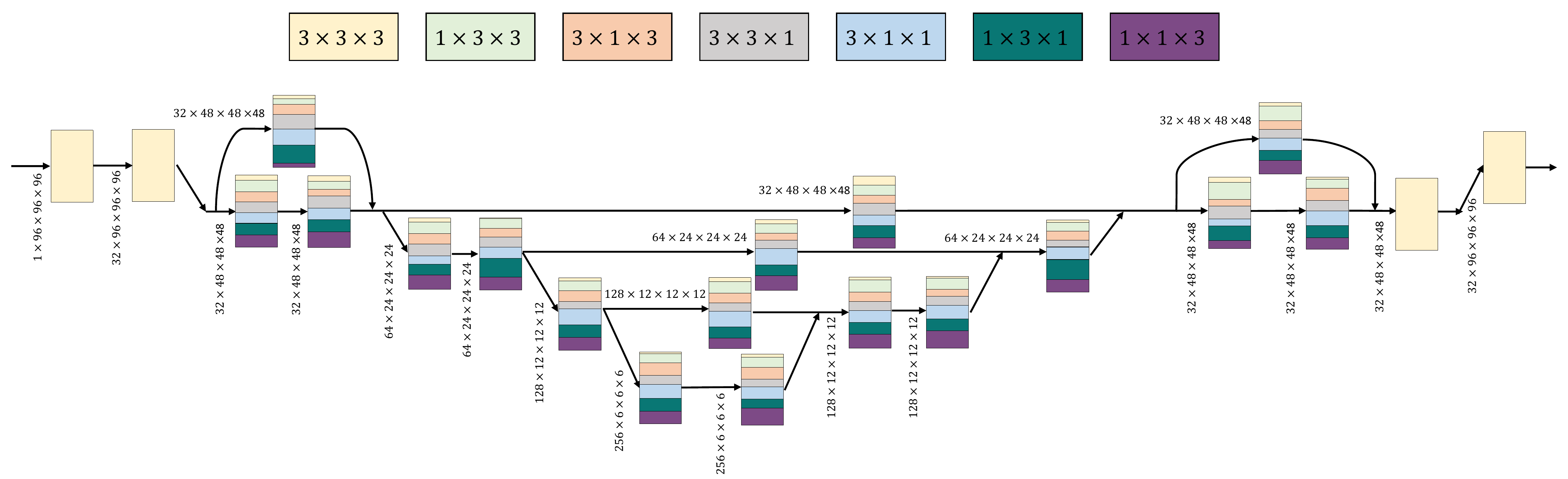}
        \caption{}
        \label{fig:fig_MSD_Arch}
    \end{subfigure}
    \\[2ex]
    \begin{subfigure}[b]{\textwidth}
        \centering
        \includegraphics[width=\textwidth]{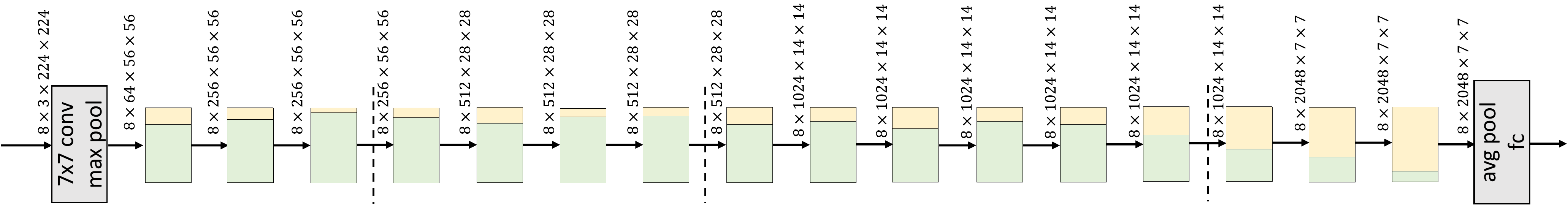}
        \caption{}
        \label{fig:fig_video_Arch}
    \end{subfigure}
    \caption{The searched architecture of $\mathrm{CAKES^{C}}$ (a) on medical data and (b) on video data. Each color represents a type of sub-kernel. The heights of these blocks are proportional to their ratios in the corresponding convolution layer. The beginning and ending $1\times 1\times1$ convolutions at each residual block are not visualized.}
\label{fig:fig_Arch}
\end{figure*}

\paragraph{Cost-Priority Architecture.} 
We plot the found architecture on both medical data ($\mathrm{CAKES^{C}_{1,2,3D}}$) and video data ($\mathrm{CAKES^{C}_{2,3D}}$) respectively in Fig~\ref{fig:fig_Arch}. \textbf{For the architecture searched on Something-Something dataset}, we note that the algorithm prefers efficient 2D operations at the bottom of the network, and favors 3D operation at the top of the network. This implies that 
the search algorithm successfully finds that \textbf{temporal information extracted from high-level features is more useful}, 
which coincides with the observation in~\cite{xie2018rethinking}. 
\textbf{For the architecture found on the MSD dataset}, we calculated the number of operations computed on all three data dimensions, and numbers are ($1285$, $1260$, $1314$). This suggests that the searched model, unlike the searched network for videos, tends to \textbf{treat each dimension equally}, which aligns with the property of volumetric medical data. 
In addition, the number of 1D, 2D, 3D operations are $1378$, $1065$, and $117$ respectively, indicating that the efficient 1D/2D operations are more preferred.

\begin{figure*}[tb]
    \begin{subfigure}[b]{\textwidth}
        \centering
        \includegraphics[width=\textwidth]{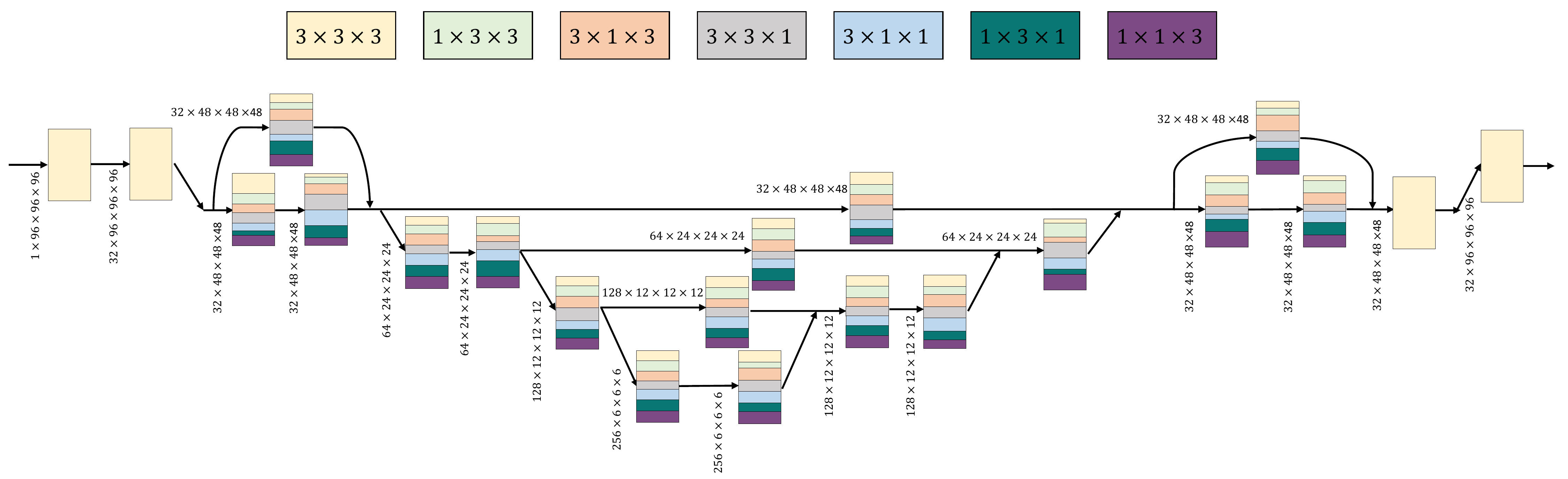}
        \caption{}
        \label{fig:fig_MSD_Arch_123D}
    \end{subfigure}
    \\[2ex]
    \begin{subfigure}[b]{\textwidth}
        \centering
        \includegraphics[width=\textwidth]{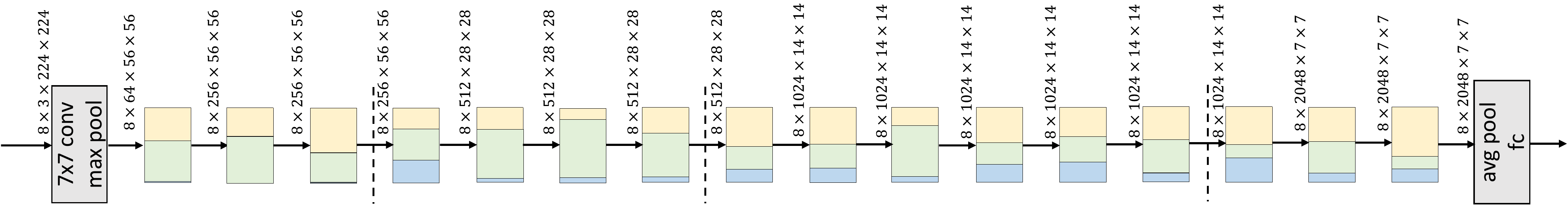}
        \caption{}
        \label{fig:fig_video_Arch_123D}
    \end{subfigure}
    \caption{The searched architecture of $\mathrm{CAKES^{P}}$ (a) on medical data and (b) on video data. Each color represents a type of sub-kernel. The heights of these blocks are proportional to their ratios in the corresponding convolution layer. The beginning and ending $1\times 1\times1$ convolutions at each residual block are not visualized.}
\label{Fig:Final_Arch_Video_123D}
\end{figure*}

\paragraph{Performance-Priority Architecture.} As shown in Fig.~\ref{fig:fig_video_Arch_123D}, for $\mathrm{CAKES^{P}_{1,2,3D}}$, the pure temporal sub-kernel ($3\times 1\times 1$) is rarely chosen at the top of the network, while it plays a more important role as the network goes deeper. This observation agrees with our previous finding: temporal information extracted from high-level features is more useful. For $\mathrm{CAKES^{P}_{1,2,3D}}$ on medical image as shown in Fig.~\ref{fig:fig_MSD_Arch_123D}, we calculate the numbers of each type of sub-kernels and computation on three axes. The numbers of 1D, 2D, and 3D sub-kernel are $1135$, $1073$, $352$ and the number of operations computed on all three data dimensions are $1437$, $1433$, $1467$, which again coincides with our previous finding that each dimension plays an equally important role for the symmetric volumetric data. Compared to cost-priority $\mathrm{CAKES^{C}}$, we notice that performance-priority $\mathrm{CAKES^{P}}$ favors the operation set with more 3D sub-kernels, which can provide a larger model capacity.
\vspace{4ex}

\end{document}